# Cross-Language Domain Adaptation for Classifying Crisis-Related Short Messages

**Muhammad Imran**
Qatar Computing Research Institute, HBKU
Doha, Qatar
mimran@qf.org.qa

**Prasenjit Mitra**
Qatar Computing Research Institute, HBKU
Doha, Qatar
pmitra@qf.org.qa

**Jaideep Srivastava**
Qatar Computing Research Institute, HBKU
Doha, Qatar
jsrivastava@qf.org.qa

**ABSTRACT**

Rapid crisis response requires real-time analysis of messages. After a disaster happens, volunteers attempt to classify tweets to determine needs, e.g., supplies, infrastructure damage, etc. Given labeled data, supervised machine learning can help classify these messages. Scarcity of labeled data causes poor performance in machine training. Can we reuse old tweets to train classifiers? How can we choose labeled tweets for training? Specifically, we study the usefulness of labeled data of past events. Do labeled tweets in different language help? We observe the performance of our classifiers trained using different combinations of training sets obtained from past disasters. We perform extensive experimentation on real crisis datasets and show that the past labels are useful when both source and target events are of the same type (e.g. both earthquakes). For similar languages (e.g., Italian and Spanish), cross-language domain adaptation was useful, however, when for different languages (e.g., Italian and English), the performance decreased.

**INTRODUCTION**

Microblogging platforms such as Twitter provide active communication channels during the onset of mass convergence events such as natural disasters (Palen et al., 2009; Hughes et al., 2009; Starbird et al., 2010; Vieweg et al., 2010). In recent years, Twitter has been used to spread news about casualties and damages, donation offers and requests, and alerts, including multimedia information such as videos and photos (Cameron et al., 2012; Imran et al., 2013a; Qu et al., 2011). Many studies show the significance of this online information (Vieweg et al., 2014; Sakaki et al., 2010; Neubig et al., 2011). Moreover, it has been observed that these messages are usually communicated more quickly than disaster information shared via traditional channels such as news websites, etc. For instance, the first tweet to report on the 2013 Westgate Mall attack was posted within a minute of the initial onslaught.[1] Given the importance of crisis-related messages for time-critical situational awareness, disaster-affected communities and professional responders may benefit from using an automatic system to extract relevant information from social media.

For rapid crisis response, real-time insights are important for emergency responders. To identify actionable and tactical informative pieces from a growing stack of social media information and to inform decision-making processes as early as possible, messages need to be processed as they arrive. Given the volume of the messages, we need to triage them. That is, we need to put them in different actionable bins such as food, supplies, financial, logistics, etc. so that disaster-response professionals can quickly look into each bin and identity the needs. Different approaches can be employed to filter and classify these messages. For instance, many humanitarian organizations use the Digital Humanitarian Network (DHN)[2] of volunteers to analyze messages one by one to find actionable information. However, given the amount of information that needs to be triaged, and the scarcity of volunteers, we would ideally like the messages to be categorized automatically and volunteers use their time to perform higher-order tasks. Despite advances in natural language processing, full automation is not feasible. Most classifiers that achieve high accuracies in solving different classification tasks are based on supervised machine learning where humans provide a set of training sample consisting of positive and negative examples for each classification category. A semi-automated system having similar characteristics to DHN is AIDR (Artificial Intelligence for Disaster Response) (Imran et al., 2014). AIDR can be trained to then automatically process and classify messages at high-speed using a supervised classification technique.

---

[1] http://www.ihub.co.ke/blog/2013/10/how-useful-is-a-tweet-a-review-of-the-first-tweets-of-the-westgate-attack

The AIDR platform collects event-specific data (using user-defined queries) from the Twitter streaming API and uses supervised machine learning techniques to classify messages into user-defined categories or bins. AIDR or other similar systems that perform automatic classification require human-labeled example messages pertaining to each category. Scarcity of labeled data results in poorer classification models. Gathering training data for such classifiers is a hard problem because human annotators find it a boring and laborious task especially if they are doing it in large numbers. However, AIDR has been used to collect data from similar events in the past and has annotated data that can be used, if they are found useful. If we can reuse the existing annotations from AIDR, then we can also improve the accuracy significantly resulting in a much better model.

In this work, we aim to utilize labels from past crises to train machines so that they can classify messages from new crises. When multiple such past crises exist, we need to choose which ones are useful and which are not. The traditional machine learning premise is that we should use as much relevant training data as we have. However, should we use labeled messages from different languages from the same event type to train? Are all the datasets from the same event, for example an earthquake, relevant for the next earthquake? Because the different datasets originate from different parts of the world, they use different languages or mix of languages, etc., the datasets for a similar event, e.g., earthquake may not be useful from one event to another. We wanted to examine the datasets to see if existing datasets and their tagged data helps.

The 800+ collections created using the AIDR system along with the human-tagged data provide a vast resource for training the classifiers. We examine the following questions empirically. 1) Can we use the past data to build models? 2) Will using the past data improve the models? 3) Should we use all the data? We train classifiers using different combinations of existing data and examine on unseen test data how they perform in order to address these questions. We show that in most cases data from the same domain are very useful. A few exceptions exist. For example, Italian tweets improved the performance of classification of tweets from Spanish-speaking countries but not English[3]. However, beyond language, there could potentially be other variations due to which we need to be careful in choosing existing data for training. For example, we believe that variations in dialects, vernaculars, season, geography, urban/rural divide, development status of countries etc. could potentially render the datasets and the discussions from the same type of event to be different.

To the best of our knowledge, our work is the first to use existing tagged information in conjunction with information tagged for the specific event to train classifiers, and show that using the old data helps improve performance in most cases. Because our evaluation found a few anomalous cases, we recommend that before deployment, we need to validate the impact of the additional datasets on the performance of the classifier using a small test set before including the training data to create the classification model. To achieve maximum performance, we should not add the training datasets that cause the classifiers to lose accuracy during this validation step.

Having established that labeled data from the same domain is generally useful, we ask the following question. Can we use data from one domain, e.g., earthquake, to train models for another domain, e.g., floods? In computer science, this is a well-known problem of domain adaptation (Daume et al., 2006). In supervised classification setting, one of the basic assumptions in learning new classifiers is that the train and test sets instances are drawn from the same data distribution. If the training and test sets differ substantially then it causes problems for the learning scheme to generalize. However, to deal with the inherited problem of labels scarcity, we aim to investigate how useful the labels from past crisis events can be for classifying a target crisis.

---

[2] http://digitalhumanitarians.com/

[3] This observation is a fascinating example of big data science. This result seems to point out that certain languages are closer to one language than others. And, that there is value in cross-language training of classification models.

The rest of the paper is organized as follows. In the next section, we describe real-time classification approach which is an application area of our current work. Datasets details and then experimental setup sections provide details regarding what datasets we use and our experimental plan. We discuss results in the discussion section and elaborate related studies in the related work section. Finally, the paper is concluded in the conclusions section.

**REAL-TIME CLASSIFICATION APPROACH**

To be useful and actionable for emergency managers during a crisis situation, information must be delivered to them in a timely fashion. In the case of social media data, this timeliness is achieved by using a real-time stream-processing paradigm (e.g. Imran et al., 2013b), in which data items are processed as soon as they arrive. Stream processing is different from batch processing, in which an archive with the information to be analyzed preexists and the processing is performed in a retrospective way.

Different data processing techniques can be used for real-time analysis of data streams (Imran et al., 2013b). In this paper we use supervised classification techniques. And, our stream processing setting involves human and machine data processing components. Specifically, humans train machines by providing labeled examples. However, human labeling cannot scale to the data volumes typical of large-scale crises, and is usually done on a sample of the input data. Whereas, automatic labeling by machines can overcome this issue, for example, by using human labeled data to train a supervised classification system. In this hybrid approach event-specific training data provided by humans is used to train and re-train an automatic classification system (e.g. Imran et al., 2014). Availability of the human labeled messages is a core aspect in this processing pipeline. However, as described earlier during the sudden onset of a crisis situation, especially in the early hours when no other means of information exist, scarcity of human-labeled data introduces a high latency to process and produce useful results for crisis responders.

To overcome this bottleneck, we study the usefulness of past labels available from previous crises. We perform extensive experimentations on a number of real crisis datasets (described next) and learn how labeled data from past crisis events can be utilized to process a new target crisis.

**DATASETS**

We use a combination of data collected by the AIDR platform and from the CrisisLexT26 dataset (Olteanu et al. 2015). Both datasets correspond to social media messages from Twitter posted during different crises that took place in 2012, 2013, and 2015. We selected 11 crises of two types: earthquake (5 crises) and floods (6 crises). Table 1 lists the crises along with other salient details. AIDR uses volunteers during the onset of a crisis situation to label crisis-related messages. However, CrisisLex used paid crowdsourcing platforms for human labeling. In the datasets, each crisis corresponds to 800+ tweets annotated using the "*Information Type*" annotation scheme, which classifies tweets into the following categories:

**Affected individuals:** deaths, injuries, missing, found, or displaced people, and/or personal updates.

**Infrastructure and utilities:** buildings, roads, utilities/services that are damaged, interrupted, restored or operational.

**Donations and volunteering:** needs, requests, or offers of money, blood, shelter, supplies, and/or services by volunteers or professionals.

**Caution and advice:** warnings issued or lifted, guidance and tips.

**Sympathy and emotional support:** thoughts, prayers, gratitude, sadness, etc.

**Other useful information:** not covered by any of the above categories.

**EXPERIMENTAL SETUP**

To determine whether labeled data from past crises can contribute to the classification of target crisis messages, we perform extensive experimentation.

**Terminologies and method:** Following are core terminologies that we use in the paper.

**Source event(s):** crisis dataset(s) used for training purposes
**Target event:** crisis dataset used for test/evaluation purposes (we always use one target event for evaluation)
**In-domain:** represents when both source and target events belong to the same crisis type (e.g. earthquake)
**Cross-domain:** represents when both source and target events have different crisis type (e.g., training on earthquakes and testing on floods)

| Crisis name (short name) | Date happened | Crisis type | Number of human labels |
|---|---|---|---:|
| **Earthquake datasets in chronological order** | | | |
| Italy earthquake (ITEQ) | 20-May-2012 | Earthquake | 911 |
| Costa Rica earthquake (CREQ) | 05-Sep-2012 | Earthquake | 866 |
| Guatemala earthquake (GUEQ) | 07-Nov-2012 | Earthquake | 905 |
| Bohol earthquake (BOEQ) | 12-Oct-2013 | Earthquake | 943 |
| Nepal earthquake (NEEQ) | 25-Apr-2015 | Earthquake | 2,812 |
| **Flood datasets in chronological order** | | | |
| Philippines floods (PHFL) | 01-Aug-2012 | Floods | 874 |
| Queensland floods (QUFL) | 29-Jan-2013 | Floods | 892 |
| Alberta floods (ABFL) | 19-Jun-2013 | Floods | 913 |
| Manila floods (MNFL) | 20-Aug-2013 | Floods | 808 |
| Colorado floods (CLFL) | 09-Sep-2013 | Floods | 901 |
| Sardinia floods (SDFL) | 17-Nov-2013 | Floods | 910 |

Table 1. Crises datasets details, their types, and number of human tagged messages

Training is always performed on data from one or more source events, and the generated models are always evaluated on one target event. The test/evaluation set remains the same for all types of experiments (more details below) for a given crisis event. The evaluation of models, especially in the domain adaptation setting, should be performed on a fixed test set, which is a more demanding evaluation task as compared to other evaluations such as cross-validation using *n*-folds.

**Preprocessing**

Preprocessing of the datasets is performed before running the experiments. Each crisis dataset is divided into two sets. The first set comprised of 70% of the messages (i.e. training set) and the second comprised of 30% of the messages (i.e. test set). For the both training and the test sets, we remove stop-words, URLs, and user mentions from the messages. We use two types of features uni-grams (one word) and bi-grams (two consecutive words). Feature selection is performed using the information gain feature selection method and top 1,000 features are selected for the training purposes. We use Random Forest, a well-known learning scheme (Liaw et al., 2002), as our classification algorithm. Results of all the experiments are presented in four well-known measures i.e. Precision, Recall, F-measure, and AUC (i.e. Area Under ROC curve).[4]

**Model adaptation using single source (in-domain and cross-domain)**

To test the performance of classifiers trained using labeled data from one event (source crisis) and test on another event (target crisis), we perform domain adaptation using single-source experiments. In this setting, we use datasets from both in-domain and cross-domains. The in-domain setting represents both train and test sets from same crisis type (e.g. earthquake). The cross-domain setting represents train and test from different crisis types.

**1- In-domain (earthquakes)**: First, we take earthquake datasets in their chronological order and use the event under investigation as target event and its preceding crises as source events. We always train classifiers on the source event data and test on the target event data. In Table 2, all the rows with experiment type "SS" represent the results obtained using the single-source experiments. For instance, the first SS row in Table 2 shows the results of training on ITEQ 100% (i.e. all Italy earthquake labels) and testing on CREQ 30% (i.e. 30% of Costa Rica earthquake labels). The Italy earthquake event happened before the Costa Rica earthquake. And the reason why Italy EQ is not tested because we don't have any preceding event to this one.

**2- In-domain (floods)**: Next, the floods datasets are tested in their chronological order. As before, the current crisis data is considered as the target event and its preceding crises the source event(s). As always, we train classifiers on source event data and test on target event data. Table 3 shows the results of in-domain (floods) experiments in rows with experiment type as "SS".

**3- Cross-domain (earthquakes and floods)**: In this setting, we performed cross-domain experiments i.e. both source and target datasets are taken from different domains. In these experiments, we aim to find out if incorporating training examples from other crisis types can increase classification accuracy or not.

Table 4 shows the results of cross-domain experiments for some selected events.

**Model adaptation using multiple sources (in-domain)**

To test whether incorporating more training examples from more than one similar past crises increases the classification accuracy or not, we perform the following two types of experiments.

**1- Using labels from more than one past crises without using any labels from target event**: More training examples tend to boost classifier's capability to generalize concepts better. To determine whether incorporating labels from all similar past crises is useful or not, in this experiment, we take all preceding datasets as our source events and used as training set. New models are trained using this training set. The evaluation of the newly generated models is performed on the test set of a target event.

Table 2 with rows having experiment type "MS" (i.e. multi-source) shows the results of all the earthquake events. Table 3 shows the results of all the floods events (rows with experiment type "MS").

**2- Using labels from more than one past crises and labels from target event**: Given the fact that classifiers generalize better if both training and test instances are drawn from the same data distribution. In this setting, we include training examples from the target event. For this purpose, we take labels (70%) from the target event to determine the boost in classification accuracy. Table 2 shows the results of earthquake events and Table 3 shows the results of floods events, both with rows having experiment type as "MSWT" (i.e. multi-source with target event).

**Model adaptation in special cases**

In supervised classification systems that make use of textual features such as uni-grams, bi-grams, or part-of-speech tags, etc., the language of the underlying data from which the features are drawn play an important role. Two events of the same type (e.g. earthquake) happened in two different countries could be effectively used to train classifiers, if the spoken language of the both countries is similar (e.g., Italian and Spanish). To determine the usefulness of such cases, in this setting, we train and test classifiers in which both source and target events are from countries where the lexical similarity between their spoken languages is high. For instance, according to the Wikipedia[5] the lexical similarity between Spanish and Italian language is almost 82%.

---

[4] https://en.wikipedia.org/wiki/Receiver_operating_characteristic

| Exp. Type | Source (s): Train set (size) | Target: Test set (size) | Precision | Recall | F-measure | AUC |
|---|---|---|---|---|---|---|
| SS | ITEQ (100%) | CREQ (30%) | 0.76 | 0.56 | 0.57 | 0.85 |
| MSWT | ITEQ (100%) + CREQ (70%) | CREQ (30%) | 0.85 | 0.85 | 0.84 | 0.95 |
| SS | CREQ (100%) | GUEQ (30%) | 0.62 | 0.55 | 0.51 | 0.85 |
| MS | ITEQ (100%) + CREQ (100%) | GUEQ (30%) | 0.77 | 0.66 | 0.69 | 0.93 |
| MSWT | ITEQ (100%) + CREQ (100%) + GUEQ (70%) | GUEQ (30%) | 0.84 | 0.85 | 0.83 | 0.97 |
| SS | GUEQ (100%) | BOEQ (30%) | 0.73 | 0.42 | 0.48 | 0.73 |
| MS | ITEQ (100%) + CREQ (100%) + GUEQ (100%) | BOEQ (30%) | 0.76 | 0.49 | 0.55 | 0.68 |
| MSWT | ITEQ (100%) + CREQ (100%) + GUEQ (100%) + BOEQ (70%) | BOEQ (30%) | 0.90 | 0.87 | 0.87 | 0.95 |
| SC1 | CREQ (100%) + GUEQ (100%) | BOEQ (30%) | 0.80 | 0.43 | 0.56 | 0.76 |
| SC2 | ITEQ-EN (100%) CREQ (100%) + GUEQ (100%) | BOEQ (30%) | 0.77 | 0.45 | 0.51 | 0.77 |
| SC3 | ITEQ-EN (100%) + CREQ (100%) + GUEQ (100%) + BOEQ (70%) | BOEQ (30%) | 0.88 | 0.85 | 0.85 | 0.97 |
| SS | BOEQ (100%) | NEEQ (30%) | 0.48 | 0.25 | 0.15 | 0.64 |
| MS | ITEQ (100%) + CREQ (100%) + GUEQ (100%) + BOEQ (100%) | NEEQ (30%) | 0.54 | 0.25 | 0.15 | 0.60 |
| MSWT | ITEQ (100%) + CREQ (100%) + GUEQ (100%) + BOEQ (100%) + NEEQ (70%) | NEEQ (30%) | 0.87 | 0.86 | 0.86 | 0.97 |
| SC1 | CREQ (100%) + GUEQ (100%) + BOEQ (100%) | NEEQ (30%) | 0.53 | 0.29 | 0.21 | 0.63 |
| SC2 | ITEQ-EN (100%) + CREQ (100%) + GUEQ (100%) + BOEQ (100%) + NEEQ (70%) | NEEQ (30%) | 0.86 | 0.86 | 0.86 | 0.98 |

**Table 2. In-domain single-source (SS), multi-source (MS), multi-source with target crisis (MSWT), and special case (SC) model adaptation results for earthquake datasets**

Rows with experiment type "SC" (i.e. special case) in Table 2 and Table 3 show the results of this analysis. For instance, in case of the Bohol Earthquake (BOEQ), we ran three additional tests. In the first test (SC1), we dropped ITEQ as it was present in the BOEQ MS case in which we observe a drop in the accuracy (e.g. see AUC).

---

[5] https://en.wikipedia.org/wiki/Lexical_similarity

However, after dropping the ITEQ, the classification accuracy increases (see SC1 row of BOEQ in Table 2). As the ITEQ set contains messages from both English and Italian languages, probably this causes the drop of AUC in the first test and the increase in AUC in the second test. To validate this observation, we manually analyzed all 912 ITEQ tweets to assign language tags (English or Italian). The result of the language tagging found that 90% of the tweets in ITEQ are in Italian language. Next, we only used ITEQ-EN (10% English set) along with CREQ and GUEQ to train a new model. The results are shown in the row with SC2 on BOEQ (30%) test set. We can see 9% increase in AUC.

| Exp. type | Source (s): Train set (size) | Target: Test set (size) | Precision | Recall | F-measure | AUC |
|---|---|---|---|---|---|---|
| SS | PHFL (100%) | QUFL (30%) | 0.60 | 0.50 | 0.51 | 0.82 |
| MSWT | PHFL (100%) + QUFL (70%) | QUFL (30%) | 0.86 | 0.85 | 0.85 | 0.97 |
| SS | QUFL (100%) | ABFL (30%) | 0.74 | 0.61 | 0.61 | 0.83 |
| MS | PHFL (100%) + QUFL (100%) | ABFL (30%) | 0.42 | 0.43 | 0.40 | 0.81 |
| MSWT | PHFL (100%) + QUFL (100%) + ABFL (70%) | ABFL (30%) | 0.80 | 0.80 | 0.79 | 0.96 |
| SS | ABFL (100%) | MNFL (30%) | 0.61 | 0.52 | 0.53 | 0.77 |
| SC1 | PHFL (100%) | MNFL (30%) | 0.70 | 0.61 | 0.60 | 0.91 |
| SC2 | PHFL (100%) + MNFL (70%) | MNFL (30%) | 0.77 | 0.75 | 0.75 | 0.95 |
| MS | PHFL (100%) + QUFL (100%) + ABFL (100%) | MNFL (30%) | 0.74 | 0.69 | 0.70 | 0.89 |
| MSWT | PHFL (100%) + QUFL (100%) + ABFL (100%) + MNFL (70%) | MNFL (30%) | 0.81 | 0.80 | 0.80 | 0.95 |
| SS | MNFL (100%) | CLFL (30%) | 0.65 | 0.54 | 0.48 | 0.85 |
| SC | QUFL (100%) + ABFL (100%) | CLFL (30%) | 0.75 | 0.67 | 0.70 | 0.94 |
| MS | PHFL (100%) + QUFL (100%) + ABFL (100%) + MNFL (100%) | CLFL (30%) | 0.80 | 0.76 | 0.76 | 0.94 |
| MSWT | PHFL (100%) + QUFL (100%) + ABFL (100%) + MNFL (100%) + CLFL (70%) | CLFL (30%) | 0.83 | 0.83 | 0.83 | 0.96 |
| SS | CLFL (100%) | SDFL (30%) | 0.55 | 0.41 | 0.29 | 0.78 |
| MS | PHFL (100%) + QUFL (100%) + ABFL (100%) + MNFL (100%) + CLFL (100%) | SDFL (30%) | 0.61 | 0.53 | 0.54 | 0.85 |
| MSWT | PHFL (100%) + QUFL (100%) + ABFL (100%) + MNFL (100%) + CLFL (100%) + SDFL (70%) | SDFL (30%) | 0.88 | 0.88 | 0.88 | 0.98 |

**Table 3.** In-domain single-source (SS), multi-source (MS), multi-source with target crisis (MSWT), and special case (SC) model adaptation results for floods datasets

In the third test, we include 70% of the BOEQ labels along with ITEQ-EN, CREQ, and GUEQ. For this the results can be seen in SC3 row of Table 2. When we use the ITEQ-EN, i.e., only the English language tweets related to the Italy earthquake, we noted an increase in the performance of new classifier.

For floods datasets, again rows with experiment type "SC" show the results of the special cases analysis. For instance, in case of MNFL, we can observe an increase in accuracy when using PHFL as train set as compared to PHFL, QUFL, and ABFL altogether for training (see rows "MS" and "SC1" of MNFL).

**DISCUSSION**

The general lesson learned from Table 2 is that including more training data, even from a mixed-language source, improves the accuracy significantly. However, the following are interesting observations.

| Source (s): Train set (size) | Target: Test set (size) | Precision | Recall | F-measure | AUC |
|---|---|---|---|---|---|
| BOEQ (100%) | PHFL (30%) | 0.38 | 0.35 | 0.26 | 0.58 |
| NEEQ (100%) | MNFL (30%) | 0.35 | 0.42 | 0.27 | 0.55 |
| MNFL (100%) | NEEQ (30%) | 0.43 | 0.31 | 0.25 | 0.59 |
| SDFL (100%) | ITEQ (30%) | 0.62 | 0.50 | 0.46 | 0.69 |
| BOEQ (100%) + MNFL (100%) | PHFL (30%) | 0.66 | 0.61 | 0.58 | 0.86 |
| BOEQ (100%) + MNFL (100%) + ABFL (100%) | PHFL (30%) | 0.64 | 0.61 | 0.58 | 0.86 |
| BOEQ (100%) + MNFL (100%) | NEEQ (30%) | 0.50 | 0.28 | 0.22 | 0.64 |

**Table 4. Cross-domain single-source model adaptation results for both earthquake and floods datasets**

1. Data from the Italy earthquake had a serious negative effect in some settings (Bohol earthquake and Nepal earthquake) but, it was useful in others (Costa Rica and Guatemala earthquakes). We believe that this exception is because 90% of the Italian earthquake data was in Italian, whereas our test case contained tweets related to earthquakes in Bohol and Nepal were primarily in English. This result seems to suggest that Italian is closer to Spanish as a language than English, an observation validated by multiple speakers of these languages and by the language-tree[6]. In cases where the language is significantly different, e.g., ITEQ versus BOEQ or NEEQ, it is better to leave the training set out. However, in these cases, it is best to select the training examples in ITEQ that are in English and using it to train in these cases, as we showed which increases the classifier performance.
2. A proposition could be made that we should segregate tweets based on language and use tweets from the same language to train and test. However, that is not an optimal strategy. Note that, for the target GUEQ, learning from the same language Costa Rica earthquake and testing it on GUEQ is worse than learning from combining the Spanish and Italian tweets. So, at least, when you do not have enough Spanish data to train, training using Italian was valuable and increased accuracy.
3. Training data from target, even in small proportion, always help increase classifiers performance. This can be seen in all experiments in which 70% of the target labels were included in the training set.

Table 3 also confirms the general philosophy that more training data is good. However, there are some interesting observations there too:

1. For the test case MNFL, using the Philippines data and the Manila data performs almost as well as when we put the other data in. This shows that using data from the same area will be immensely useful because the language mixture (Tagalog, English) used in these two cases is almost exactly the same. However, adding QUFL and ABFL, which are solely in English, seems to improve the performance slightly.

Generally, we see that tagging a few tweets related to the same earthquake still improves the performance significantly. Perhaps this may mean that we still do not have enough data and in the future, when we collect more data, we can eliminate the requirement for training on the current (target) dataset.

---

[6] An interesting by-product of our work could be to construct a language-tree and language-language distances based on online language in disaster-related tweets. Perhaps such a tree/distance measure could be then used to select which languages can be used for cross-training and which should not be used, especially in cases where we have few training examples in one language.

Initial results are promising to show that there may be some signal in using the flood related tweets to augment earthquake tweets but it also has the chance of reducing the accuracy of the classifier. For example, Table 4 (last row) shows that adding MNFL to BOEQ increased the performance on the test set NEEQ. However the previous two rows show a slight decrease in accuracy by adding the flood-related training set. The general consensus seems to be that given our collection of tagged tweets from the past, we should stick to using all the tweets from the same domain provided the language mixture is similar. At this point, using cross-domain training sets have not conclusively shown any consistent improvement in the accuracy.

**RELATED WORK**

Mass convergence events, particularly those with no prior warning, require rapid analysis of the available information to make timely decisions. Information posted on microblogging platforms during crises can aid crisis response efforts, if processed timely and correctly (Yin et al. 2012; Starbird et al., 2010; Palen et al., 2009). Many approaches based on human annotation, supervised learning, and unsupervised learning techniques have been proposed to process social media data---for a complete survey see e.g., (Imran et al., 2015).

In this work, we use supervised machine learning techniques to classify crisis-related messages (many such efforts and systems based on these techniques have been developed in past e.g., (Mendoza et al., 2010; Olteanu 2015; MacEachren et al., 2011; Imran et al., 2014; Roy et al., 2013)). For instance, ESA (Yin et al. 2012; Power et al. 2014) uses naıve Bayes and SVM, EMERSE (Caragea et al., 2011) uses SVM, AIDR (Imran et al. 2014) uses random forests, and Tweedr (Ashktorab et al. 2014) uses logistic regression. However, due to the scarcity of training data, which is one of the basic ingredients for such approaches to work well, causes delays in machine training.

To overcome the issue of scarcity of the training data for a new crisis, we study the usefulness of labels (training data) from past crises. Li et al., (2015) studied the problem of domain adaptation. They combine source labeled data with target unlabeled data to train classifiers (Naive Bayes in their case) and observed a high performance by including target crisis data in training set as compared to only source crisis data. Their findings, to some extent, are inline with ours; however, the evaluation mechanism that they have used is based on cross-validation using 5-fold setting. However, in our case, we always use a holdout test set across all variations of experiments, which is a more challenging problem in an online classification setting. Moreover, we also provide empirical results by training models in cross-language settings.

**CONCLUSIONS**

Availability of training data to train machine learning classifiers during the early hours of a crisis situation can help gain early insights for rapid crisis response. We show that using labeled data from past events of the same type are generally always useful if the training and testing data are from the same language. When there are not enough tweets in the one language (e.g., Spanish), labeled tweets in a different language (e.g., Italian) can be useful if the two languages in question are very similar (e.g., Italian and Spanish) but not when they are not (e.g., Italian and English/Tagalog). If there are reasonable number of labeled tweets from the same domain (e.g., earthquakes), then, we could not establish the utility of using labeled tweets from a different domain (e.g., floods). In one such case, the performance improved slightly while in another it decreased. Further systematic evaluation on these lines is needed.